\documentclass{article}
\usepackage[backend=biber,style=numeric,sorting=nyt]{biblatex} 
\usepackage{booktabs}
\usepackage{colortbl}
\usepackage{longtable}
\usepackage[most]{tcolorbox}
\definecolor{primarycolor}{HTML}{C9E4CA}
\definecolor{secondarycolor}{HTML}{F7F7F7}
\definecolor{thirdcolor}{HTML}{F5F5DC}
\newtheorem{researchq}{Research Question}
\newtcolorbox{defbox}[3][]{
fontupper=\itshape,
arc=5mm,
lower separated=false,
fonttitle=\bfseries,
colbacktitle=black!10,
coltitle=black!50!black,
enhanced,
attach boxed title to top left={xshift=0.5cm,
        yshift=-2mm},
colframe=black!50!black,
colback=gray!10}
\usepackage{arxiv}
\usepackage[utf8]{inputenc} 
\usepackage[T1]{fontenc}    
\usepackage{hyperref}       
\usepackage{url}            
\usepackage{booktabs}       
\usepackage{amsfonts}       
\usepackage{nicefrac}       
\usepackage{microtype}      
\usepackage{lipsum}		
\usepackage{graphicx}
\usepackage{doi}

\title{Human-Centered Evaluation of RAG Outputs: A Framework and Questionnaire for Human–AI Collaboration}

\author{ \href{https://orcid.org/0009-0006-8240-0922}{\includegraphics[scale=0.06]{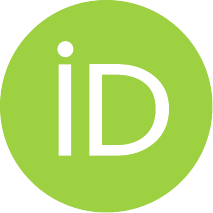}\hspace{1mm}Aline ~Mangold} \\
	Department Speculative Transformation\\
	Dresden University of Technology\\
	Dresden, 01062 \\
	\texttt{aline.mangold@tu-dresden.de} \\
	\And
	\href{https://orcid.org/0009-0003-5329-599X}{\includegraphics[scale=0.06]{orcid.pdf}\hspace{1mm} Kiran ~Hoffmann} \\
	Department Speculative Transformation\\
	Dresden University of Technology\\
	Dresden, 01062 \\
	\texttt{kiran.hoffmann@tu-dresden.de} \\
}

\renewcommand{\shorttitle}{\textit{arXiv} Template}

\hypersetup{
pdftitle={A template for the arxiv style},
pdfsubject={cs},
pdfauthor={Aline ~Mangold, Kiran ~Hoffmann},
pdfkeywords={LLM, Human-centered Evaluation, Human-AI collaboration, RAG},
}

\begin{document}
\maketitle
	\begin{abstract}
    Retrieval‑augmented generation (RAG) systems are increasingly deployed in user‑facing applications, yet systematic, human-centered evaluation of their outputs remains underexplored. Building on Gienapp’s utility‑dimension framework, we designed a human‑centred questionnaire that assesses RAG outputs across 12 dimensions. We iteratively refined the questionnaire through several rounds of ratings on a set of query‑output pairs and semantic discussions. Ultimately, we incorporated feedback from both a human rater and a human‑LLM pair. Results indicate that while large language models (LLMs) reliably focus on metric descriptions and scale labels, they exhibit weaknesses in detecting textual format variations. Humans struggled to focus strictly on metric descriptions and labels. LLM ratings and explanations were viewed as a helpful support, but numeric LLM and human ratings lacked agreement. The final questionnaire extends the initial framework by focusing on user intent, text structuring, and information verifiability.
	\end{abstract}

	
	
	
	\maketitle
	
\section{Introduction}
Due to their impressive generative capabilities \cite{schillaciLLMAdoptionTrends2024}, large language models (LLMs) have become widely adopted across various domains, including healthcare \cite{liuReviewApplyingLarge2025}, education \cite{razaIndustrialApplicationsLarge2025}, and customer service \cite{meduriRevolutionizingCustomerService2024}. However, despite their scale, these models suffer from critical shortcomings, including hallucinations (factually incorrect outputs), outdated knowledge, and a lack of domain-specific expertise \cite{wuRetrievalAugmentedGenerationNatural2025}.
\par
Retrieval-Augmented Generation (RAG) has emerged as a solution by allowing LLMs to retrieve relevant, external documents from dynamic databases and ground their responses in this evidence \cite{lewisRetrievalAugmentedGenerationKnowledgeIntensive2020}. This architecture significantly enhances factual accuracy and domain specificity, especially in knowledge-intensive tasks \cite{yangRetrievalAugmentedGenerationQuantized2024}. With an increasing adoption, RAG Systems help to extract and aggregate information from company- or domain-specific knowledge bases \cite{schillaciLLMAdoptionTrends2024}.
However, RAG systems have been observed to exhibit various shortcomings, including diminished answer quality when diluted with irrelevant documents or hallucinations if the link between input (retrieved documents) and output (generated answer) is not properly aligned \cite{guptaComprehensiveSurveyRetrievalAugmented2024}. To diagnose and address such issues, it is necessary to evaluate RAG outputs and incorporate the results into further development.
\par
Prior work has already proposed various RAG evaluation metrics, such as relevance, accuracy, faithfulness and correctness \cite{yuEvaluationRetrievalAugmentedGeneration2025}. For instance, \cite{papadimitriouRAGPlaygroundFramework2024}, distinguish programmatic metrics and LLM-based metrics. Programmatic metrics ensure lexical and semantic accuracy, e.g., the preservation of domain-specific terms or ensuring an adequate percentage of ground truth tokens. LLM-based metrics, on the other hand, capture more nuanced aspects of answer quality, such as the completeness of relevant aspects in the response.
RAG evaluation metrics can be calculated quantitatively. For instance, accuracy is computed by estimating the proportion of true results (true positives and true negatives) among the total number of cases examined. Evaluation metrics in other frameworks also employ the "LLM as a judge" method. For instance, "context relevance" can be evaluated by an LLM \cite{yuEvaluationRetrievalAugmentedGeneration2025}.
However, little attention has been paid to the development and evaluation of RAG outputs from a human-centered perspective, e.g., by employing human judges \cite{yuEvaluationRetrievalAugmentedGeneration2025}. This might lead to serious usability concerns. As computer-centered RAG metrics primarily ensure the factual accuracy and relevance of the output, crucial aspects such as optimal answer formatting, argumentation structure, or match between user intent and system output have been neglected in past research. Furthermore, while the LLM as a judge and human as a judge methods both have been explored, there is a lack of research on how machine and human evaluations can be integrated collaboratively.
To address this gap, we propose the following research questions:
\begin{researchq}
Which metrics are relevant for a human-centered evaluation of RAG outputs?
\end{researchq}
\begin{researchq}
How can relevant metrics be integrated into a questionnaire for a human-centered evaluation \noindent of RAG outputs?
\end{researchq}
\begin{researchq}
How can humans and LLMs work collaboratively in the human-centered evaluation of RAG outputs?
\end{researchq}
In this paper, we will further provide a framework of text-based RAG output evaluation grounded on prior research. We also present anchored items to enable framework usage as a questionnaire. Further, we provide an approach to integrate an LLM as an additional judge into the evaluation process. Lastly, we evaluate our questionnaire with two independent raters and refine the questionnaire based on their feedback.
\section{Prior Work and Motivation}  
\subsection{RAG System Applications and Limitations}
To this date, LLMs are adopted by organizations to solve classical natural language processing (NLP) tasks like question-answering, summarization, machine translation or text generation \cite{schillaciLLMAdoptionTrends2024}. Further applications include programming support, customer support using personalized chatbots, and conducting specialist tasks like data analysis as an AI agent. Another rising field is the integration of AI into knowledge management and information retrieval. Connecting specific knowledge bases to LLMs is usually done with a RAG pipeline. The system first pulls relevant entries from the knowledge base, then feeds those entries together with the user’s query to the LLM. This gives the model up‑to‑date, factual context, steering its output and reducing hallucinations. Research indicates that RAG systems indeed help to overcome LLM problems like low accuracy or hallucinations, for instance, in the health \cite{alyDevelopmentValidationPediatric2024} or legal domain \cite{pipitoneLegalBenchRAGBenchmarkRetrievalAugmented2024}.
\par
As RAG systems are implemented in various domains, knowledge base data may comprise a variety of modalities extending beyond text. For this reason, there are multi-modal, video-based, and audio-based RAG models \cite{guptaComprehensiveSurveyRetrievalAugmented2024}. However, to date, text-based RAG models represent the most mature and widely researched category. These systems are designed to facilitate retrieval and generation tasks, including question answering and summarization.
\par
While RAG models offer improved factuality by grounding generation in external sources, they face key challenges. Retrievers often struggle with ambiguous or niche queries, leading to irrelevant results. Additionally, poor integration between retrieval and generation can cause inconsistencies or incoherence between the retrieved text and the answer. More precisely, this is the case if a retrieval module fetches a passage and the generator blends it into an unrelated response \cite{guptaComprehensiveSurveyRetrievalAugmented2024}.
To identify and potentially address these issues in future development, RAG systems must be evaluated. As outlined in the literature, a variety of evaluation frameworks have been introduced \cite{papadimitriouRAGPlaygroundFramework2024, yuEvaluationRetrievalAugmentedGeneration2025}. In this paper, we distinguish between \textit{computer-centered} evaluation metrics and \textit{human-centered} evaluation metrics. In our context, computer-centered metrics refer to metrics that can be solely evaluated by machines and/or quantitatively calculated. They are targeted at improving system performance metrics like accuracy, relevance, or speed. Human-centered metrics can be either evaluated by LLMs, humans, or in a human-LLM collaboration. They are primarily aimed at enhancing human satisfaction, understanding, and usability of the output. Note that the distinction is not always possible without ambiguity. For instance, enhanced relevance between the user query and the retrieved information could lead to improved human satisfaction. However, it can be assessed both by calculating similarity metrics (e.g., Euclidean distance) or by subjective ratings. 
\subsection{Computer-centered RAG evaluation}. \cite{papadimitriouRAGPlaygroundFramework2024} proposed a framework for systematic evaluation of retrieval strategies in RAG systems. Their metrics are categorized into programmatic, LLM-based, and hybrid types. Programmatic metrics, such as "key terms precision" and "token recall", assess lexical and semantic accuracy by measuring the presence of domain-specific vocabulary and the completeness of information relative to the ground truth without requiring LLM intervention. LLM-based metrics, including "truthfulness", "completeness", "source relevance", and "context faithfulness", leverage structured prompting to evaluate factual accuracy, coverage of key points, and the faithfulness of responses to retrieved context, which is critical for detecting hallucinations and unsupported inferences. Hybrid metrics like "semantic F1" and "answer relevance" combine embedding-based similarity with LLM judgment to balance computational efficiency and semantic understanding, allowing for nuanced assessment of paraphrased or semantically equivalent responses.
\par
\cite{yuEvaluationRetrievalAugmentedGeneration2025} categorised evaluation metrics into two main groups: those assessing the retrieval component and those assessing the generation component. Retrieval metrics focused on how effectively the system identified and selected relevant information, considering dimensions such as "relevance" (the degree to which retrieved content addresses the user’s query), "accuracy" (the correctness and precision of the retrieved results), "diversity" (the extent to which the results represent a broad range of perspectives or sources), and "robustness" (the consistency of retrieval performance across varying queries and conditions). In contrast, generation metrics evaluated the quality of the system’s produced responses, emphasising "faithfulness" (the extent to which responses remain grounded in the retrieved evidence), "factual correctness" (the accuracy of the information conveyed), and "coherence" (the logical flow and readability of the response). Metrics were further categorised by the raw target metric ("What is evaluated?") and metric generation ("How is the metric evaluation generated?"). For instance, "context relevance" evaluation was generated by an LLM ("LLM as a Judge") in a reviewed study \cite{esRagasAutomatedEvaluation2025}, while consistency evaluation was generated by humans ("Human evaluation") in another study \cite{wangFeB4RAGEvaluatingFederated2024}.
However, \cite{yuEvaluationRetrievalAugmentedGeneration2025} noticed that LLM output evaluations might not align with human judgements. For instance, criteria for "correctness", "clarity", and "richness" could differ between LLM and human judgements. This is in line with other research, indicating that, especially for settings that require specialised knowledge, such as dietetics and mental health, agreement is rather low \cite{szymanskiLimitationsLLMasaJudgeApproach2024}. Furthermore, the authors identified a lack of a universally applicable grading scale and prompting text, which complicates the standardisation of "LLM as a Judge". In conclusion, there is a clear need for human-centred metrics that can be evaluated using a standardised grading system \cite{yuEvaluationRetrievalAugmentedGeneration2025}.
\subsection{Human-centered RAG evaluation}
To the best of our knowledge, human-centered RAG evaluation frameworks are rare. For instance, \cite{ikeAutomatingDialogueEvaluation2025} compared human and AI assessments across various dialogue scenarios. Thereby, they focused on seven key performance indicators (KPIs): "Coherence", "Innovation", "Concreteness", "Goal Contribution", "Commonsense", "Contradiction", "Incorrect Fact", and "Redundancy". Although the authors identified strong potential for LLMs to automate dialogue evaluation due to their robust capabilities in detecting factual accuracy and commonsense understanding, a core limitation was the weak correlation between human and LLM judgements. For this reason, the authors demand a combined framework that incorporates both human and LLM judgements. Another shortcoming of the aforementioned KPIs in a RAG context is that they are only partially adequate for RAG tasks. For example, the "innovativeness" KPI refers to the extent to which the output provides new and creative ideas or solutions. However, this metric is inappropriate for RAG contexts since tasks mostly involve knowledge extraction and summarisation based on existing specialised databases. Furthermore, the rating system comprised a scoring system between 0 to 100\%. This might be insufficient to properly distinguish between scores, since there are no content-related anchors included. More specifically, pure numeric values might not be sufficient to distinguish between different metric levels. For instance, an incorrect fact score of 50\% could mean that only parts of a statement are true, but it could also mean that specific details of the claim are false, while the generic claim is correct.
\par
\begin{figure}[h]
  \centering
  \includegraphics[width=11cm]{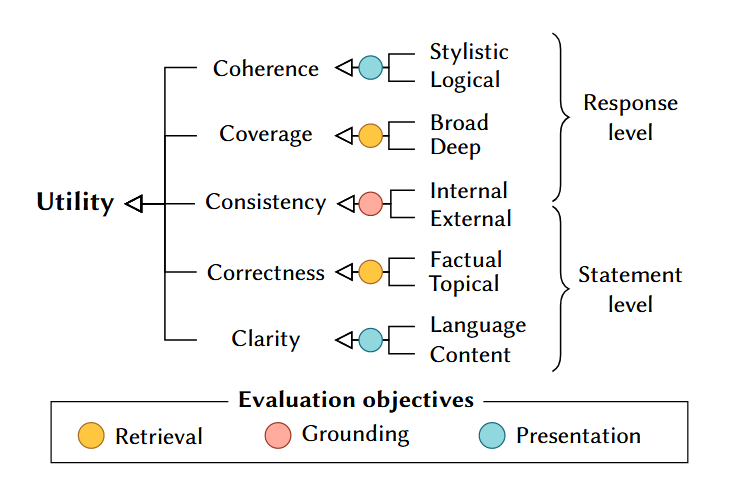}
  \caption{Taxnonomy of utility dimensions \cite{gienappEvaluatingGenerativeAd2024}. Licensed under CCAI 4.0 License.}
  \label{pic:utility_taxonomy}
\end{figure}

\cite{gienappEvaluatingGenerativeAd2024} proposed a taxonomy of utility metrics categorized by evaluation objectives in information retrieval systems. The evaluation objectives include the processes of retrieval, grounding, and presentation. For instance, "presentation" refers to the model's ability to condense relevant information from multiple sources into a single answer. The concept is reflected in a range of metrics, including "stylistic coherence", "logical coherence", "language clarity", and "content clarity". The first ones ensure that a narrative without contradictions is formed. The latter ones relate to the presentation of statements in a clear and comprehensible manner. "Grounding" refers to the model's ability to point to source documents as evidence in the response generation process. In this regard, "internal consistency" relates to the degree of consistency within statements, while "external consistency" describes the level of consistency between a statement and the source documents. Finally, "retrieval"  is the system's capacity to identify relevant, diverse, informative, and accurate sources from a given collection. In this regard, "deep coverage" and "broad coverage" describe the diversity and depth of the included information. Lastly, "Factual Correctness" and "Topical Correctness" belong to the retrieval objective. Factual correctness captures if a statement is objectively true, and topical correctness, if it aligns with a user's informational need.
For an overview of the taxonomy, please refer to  figure \ref{pic:utility_taxonomy}.
\par
The proposed taxonomy provides accuracy and completeness metrics, as well as metrics aimed at human perception and understanding. This makes it a solid foundation for a human-centered evaluation framework. However, to the best of our knowledge, the taxonomy has not yet been applied to real RAG outputs and further developed. Furthermore, there is a lack of guidance on whether utility metrics should be assessed by a human, an LLM, or both working together. To apply the taxonomy to real data, it must be transformed into a questionnaire format. Each item should receive a rating scale with content-related anchors.
\section{Methods}
\subsection{Questionnaire Development}
\subsubsection{Selected Method}
For questionnaire development, we chose a method similar to directed content analysis \cite{hsiehThreeApproachesQualitative2005}. Directed content analysis is targeted at expanding existing theoretical frameworks. However, in this context, the extension was not related to the development of codes but rather to refining and adding utility metrics to the existing framework.
\subsubsection{Initial Questionnaire}
As a first step to adapt the taxonomy by \cite{gienappEvaluatingGenerativeAd2024} into a questionnaire, we generated items, item descriptions, the establishment of a rating scale, and content-related labels for the scale. Research indicates that labeling responses can increase reliability in questionnaires \cite{wengImpactNumberResponse2004}. In the first draft, each utility metric (e.g., "stylistic coherence") represented an item. For each item, the response format consisted of a 5-point Likert scale. The scale values 1, 3, and 5 received textual labels. For certain items, a no-answer category was established.
\subsubsection{First Application}
In the next step, we used the draft on real RAG output data. Data was generated in a user test, which was part of another study. In these tests, five researchers questioned a RAG system regarding either one research document (single-document chat) or two research documents (multi-document chat). User queries were related to content extraction, such as "Summarize the paper" and "What are the key points of the paper?" This resulted in a total of 38 query-output pairs. In an initial evaluation round, we rated the outputs with the questionnaire draft. In discussions, we identified various shortcomings of the questionnaire. One major issue was the lack of consistent wording across different scale values in the same items. Since the included aspects were not uniformly distributed, this led to confusion with numeric ratings. Another problem was the semantic separability of the metrics. For instance, "logical coherence" and "internal consistency" were closely related semantically. Furthermore, several aspects were missing. For instance, some outputs repeated the same aspects multiple times. This redundancy problem was not reflected in any of the included metrics. For that reason, we revised the questionnaire items, descriptions, and scale labels.
\subsubsection{Refinement}
Following several iterations of semantic discussions, we applied a revised version of the questionnaire another time to a subset of the aforementioned query-output pairs. We identified further shortcomings, including aspects such as "salient clarity" (the degree to which key information is clearly presented in the textual response) and "model clarity" (the presence of an explanatory element that accompanies the generation of outputs). These address the issue of explainability in LLM-generated outputs. We further refined item descriptions and item labels to match a uniform language style. 
\subsubsection{Final Questionnaire}
Ultimately, the questionnaire consisted of 12 metrics, using a 5-point Likert scale. Some scales included a no-answer option. For instance, "broad coverage" could only be assessed if the user query asked for a diverse point of view. To prepare the questionnaire for a human and LLM-collaborated evaluation, we divided the items depending on whether they should be solely rated by a human or rated by a human-LLM collaboration. We selected items that could be rated by only reviewing the query output pairs for a human-LLM collaboration, for instance, "logical coherence". We rated items, which incorporated verification work like fact-checking in the source document, as solely human.
\subsection{Incorporation of LLM as a Judge}
To generate LLM ratings, we wrote a script in Python 3.12, using GPT-4o as an LLM. We evaluated the metrics categorized as suitable for a human-LLM collaboration, using a subset of the aforementioned query-output pairs. Initially, we found that the LLM did not focus solely on the relevant metrics in the final ratings. For this reason, we specified this focus in the system prompt and evaluated each metric with a separate API call.
The final system prompt was the following:
\begin{defbox}{}{}
"You are an impartial evaluator. Your task is to assess AI-generated answers about research papers to user prompts based on specific criteria. You will receive:
1. The original user prompt.
2. The AI-generated output.
3. An evaluation criterion and its description.
Based only on the provided criterion and its description, rate the output on a scale from 1 to 5. Do not use a criterion other than the provided criterion and its description for rating. Explain your rating in one sentence, referencing specific parts of the output as justification." 
\end{defbox}{}{}
The final user prompt was the following:  
\begin{defbox}{}{}
"Please rate the output \{output\} which is related to the prompt \{user\_prompt\}. Use the \{criterion\_information\} which is related to \{evaluation\_criterion\} for your rating.
Provide one sentence of explanation with an example from the text, why you chose this rating."
\end{defbox}
To evaluate the questionnaire and the human-LLM collaboration process on unseen data, we had to generate new query-output pairs. We generated a new set of four extraction-related queries (e.g., "From where and how were the study participants recruited?") and applied this to a new research document using GPT-OSS-120B.
\subsection{Final evaluation}
For the final evaluation, we established two settings:
\begin{enumerate}
    \item \textbf{Collaborative Human-LLM evaluation}: Rating query-output pairs with the questionnaire, metric descriptions, research document, and additional ratings and explanations of an LLM.
    \item \textbf{Human evaluation}: Rating query-output pairs with the questionnaire metric descriptions and the research document.
    \end{enumerate}
Before providing their ratings, we gave the raters textual instructions explaining what RAG systems are and how the outputs were generated. They were encouraged to read the research document and the associated descriptions of the metrics before rating. After rating, they filled out a survey, rating the metrics regarding their understandability ("The dimension was understandable to me"; \textit{completely disagree/completely agree}) and usefulness ("The dimension was useful for me"; \textit{completely disagree/completely agree}). They could also provide suggestions for improvement for each metric and propose additional metrics that were missing. The rater in the human-LLM condition further assessed the LLM ratings and explanations regarding their agreement with them and their usefulness. They were also asked to describe situations in which they disagreed with the LLM.
Overall, two raters participated in the process, each assigned to one condition. The whole process lasted about 2.5 hours per rater.
After the survey condition, the intraclass correlation between the two raters was calculated.
\section{Results}
\subsection{Ratings}
The rating section consisted of the numerical ratings given by the raters. Additionally, the rater in the human-LLM evaluation provided qualitative feedback to support their ratings. For both human raters, ICC was calculated. For human-LLM, no ICC was calculated, since ratings were not independent.
\subsubsection{Rater Agreement}
In this context, the raters were both selected at random. Each query-output pair represents a subject. In total, four subjects were evaluated on twelve items. Conclusively, ICC(2,1) was selected as the appropriate convention. The results indicated an ICC(2,1) value of .73, 95\% CI [.30, .91], F(11, 11) = 6.16, p = .003. This suggests a good level of reliability and a significant level of agreement between the raters.
\subsubsection{Human-AI Collaboration}
There were several differences between human and LLM ratings. Concerning "stylistic coherence", outputs received a high LLM rating if they contained sufficient formatting, although some formatting styles, for instance, using (a), (b), and (c) as counting marks, were not recognized by the LLM. The human rater was not satisfied with the existing formatting and emphasized in most cases that it lacked consistency (e.g., formatting styles were mixed), although this was not demanded in the metric description. Another difference in ratings was found in the "topical correctness" metric. While the LLM accurately attributed ratings to user intent, human ratings were lower if the information was factually incorrect. This suggests a lack of clarity regarding the term "topical correctness". Ratings regarding "logical coherence", "language correctness", and "saliency clarity" were similar between human and LLM.
\subsection{Survey} 
The survey consisted of a quantitative part (ratings on a Likert scale), and a qualitative part (suggestions for improvement, suggestions for additional metrics). Due to the small sample size, no inferential statistics were conducted.
\subsubsection{Quantitative}
The mean rating for most metrics ranged from 4 to 5 (calculated by aggregating the ratings from the two raters and the understandability and usefulness items). "Broad coverage" was an exception, receiving an aggregated rating of 3.5. The overall questionnaire received an aggregated rating of MW = 4.60 and SD = 0.49. The rater receiving the LLM ratings evaluated their agreement with the ratings as partially congruent (3.0) and the LLM ratings as useful (5.0). The LLM justifications were rated as rather logical (4.0) but only partially correct (3.0).
\subsubsection{Qualitative}
Raters could provide open-ended feedback on each metric. Both parties expressed confusion regarding the "broad coverage" metric. Firstly, there was an incongruence between the metric description and the scale labels. Additionally, the definition of "diverse information" was ambiguous. One of the raters mentioned that it was not feasible to support diverse points of view, since there was only one source document included. The other rater suggested that the metric should refer to "completeness" rather than diversity. Concerning the concept of "factual correctness", one rater noted that the definition of "objectively true" could be ambiguous, as it could refer to the document itself, the rater's logical thinking, common sense, or the need for further verification, such as through web research. The third metric that was the subject of criticism was "model clarity". This referred to the provision of an explanation alongside the model output. One rater suggested that explanations could be links to the source document, but that, for them, an explanation would consist of a causal component. They emphasised that such an explanation would not be adequate in this context.
\subsubsection{Questionnaire}
After incorporating rater feedback, the final questionnaire consists of 12 metrics, each evaluated on a 5-point Likert scale. Four of them are categorized as human evaluation only ("broad coverage", "deep coverage", "external consistency", and "verifiability correctness") because they involve fact-checking in the source document.
The metric "model clarity" was discarded, as it was deemed infeasible to include understandable textual explanations for every output generation. We removed "factual correctness" because ensuring the output is consistent with the source document is already covered in "external consistency". 
Further, we reworded the metric "broad coverage", aiming at output completeness rather than diversity. We also reworded "Stylistic coherence", now referring to a consistent use of formatting styles. Lastly, we renamed "topical correctness" to "user intent correctness" to emphasize the match between user intent and output.  Consistent language use, on the other hand, is now reflected in a new metric, "language consistency". We added "verifiability correctness" as a new metric, describing the extent to which provided statements are verifiable in the source text.
For an overview of the questionnaire, please refer to table ~\ref{questionnaire_table}.
\begin{longtable}{p{2cm}p{1cm}p{4cm}p{2cm}p{4cm}}
\caption{Questionnaire table. Green colour indicates human‑LLM evaluation. Grey colour indicates human only evaluation.}
\label{questionnaire_table}
\\
\hline
\multicolumn{2}{l}{\textbf{Metric}} & \textbf{Description} & \textbf{Scale} & \textbf{Scale Label} \\
\hline
\rowcolor{thirdcolor}
\multicolumn{5}{l}{\textbf{Coherence}} \\
\rowcolor{primarycolor}
& \textbf{Logical} &
Response forms a narrative without contradictions.E.g. “Is the response well‑structured in terms of content?” &
\begin{tabular}{lp{5.7cm}}
5 & clearly structured line of reasoning; individual statements are understandable and without contradictions \\
4 &  \\
3 & partly structured line of reasoning; individual statements are confusing or contradicting \\
2 &  \\
1 & unstructured/missing line of reasoning; confusing or contradicting statements
\end{tabular} \\
\rowcolor{primarycolor}
& \textbf{Stylistic} &
Consistent use of visual text formatting.E.g. “Is highlighting or bullet points used consistently?” &
\begin{tabular}{lp{5.7cm}}
5 & consistent use of formatting \\
4 &  \\
3 & some consistent use of formatting \\
2 &  \\
1 & no consistent use of formatting\\
&\\
&
\end{tabular} \\
\hline
\rowcolor{thirdcolor}
\multicolumn{5}{l}{\textbf{Coverage}} \\
\rowcolor{secondarycolor}
& \textbf{Broad} &
How well does a response treat a user's contextual information need? Does it cover relevant aspects regarding the given source and query? &
\begin{tabular}{lp{5.7cm}}
5 & all aspects relevant to the user query have been covered \\
4 &  \\
3 & some aspects relevant to the user query have been covered \\
2 &  \\
1 & no aspects relevant to the user query have been covered\\
&
\end{tabular} \\
\rowcolor{secondarycolor}
& \textbf{Deep} &
How well does a response treat a user's contextual information need? Does it provide in‑depth and highly informative content regarding the given source and query? &
\begin{tabular}{lp{5.7cm}}
5 & depth of information fits the query \\
4 &  \\
3 & a little too deep or too shallow for the query \\
2 &  \\
1 & too deep or too shallow for the query\\
&\\
&\\
&\\
&\\
&\\
&
\end{tabular} \\
\hline
\rowcolor{thirdcolor}
\multicolumn{5}{l}{\textbf{Consistency}} \\
\rowcolor{secondarycolor}
& \textbf{External} &
Consistency between a given source and the produced output &
\begin{tabular}{lp{5.7cm}}
5 & word‑by‑word consistency / absolutely correct summary \\
4 &  \\
3 & no word‑by‑word consistency / meaning remains the same \\
2 &  \\
1 & differences between source and response in terms of meaning / hallucinations
\end{tabular} \\
\rowcolor{primarycolor}
& \textbf{Lang-uage} &
Consistent language use. E.g., “Does the writing tone stay consistent?” &
\begin{tabular}{lp{5.7cm}}
5 & consistent language use \\
4 &  \\
3 & partially consistent language use \\
2 &  \\
1 & inconsistent language use
\end{tabular} \\
\hline
\rowcolor{thirdcolor}
\multicolumn{5}{l}{\textbf{Correctness}} \\
\rowcolor{secondarycolor}
& \textbf{Verifia-bility} &
Information is verifiable in the source. &
\begin{tabular}{lp{5.7cm}}
5 & information is easily verifiable in the source \\
4 &  \\
3 & some information is verifiable in the source, some is not \\
2 &  \\
1 & information is not verifiable in the source.
\end{tabular} \\
\rowcolor{primarycolor}
& \textbf{User intent} &
Information aligns with user’s thematic information need (statement fits user intent) &
\begin{tabular}{lp{5.7cm}}
5 & response fits the thematic context of the query \\
4 &  \\
3 & response fits the thematic context of the query partially \\
2 &  \\
1 & response fails to fit the thematic context of the user’s needs/query
\end{tabular} \\
\rowcolor{primarycolor}
& \textbf{Lang-uage} &
Lexically and grammatically correct language &
\begin{tabular}{lp{5.7cm}}
5 & completely correct language (lexically and grammatically) \\
4 &  \\
3 & partially correct language (lexically and grammatically) \\
2 &  \\
1 & incorrect language (lexically and grammatically)
\end{tabular} & \\
\hline
\rowcolor{thirdcolor}
\multicolumn{5}{l}{\textbf{Clarity}} \\
\rowcolor{primarycolor}
& \textbf{Lang-uage} &
Concise, comprehensible, user‑accessible and well‑spoken language &
\begin{tabular}{lp{5.7cm}}
5 & clear language / meeting conversation style \\
4 &  \\
3 & unclarity / missing conversation style partially \\
2 &  \\
1 & unclear language / missing conversation style
\end{tabular} \\
\rowcolor{primarycolor}
& \textbf{Saliency} &
How well key information is placed within the textual response &
\begin{tabular}{lp{5.7cm}}
5 & key information is completely salient \\
4 &  \\
3 & key information is partially salient \\
2 &  \\
1 & key information is not salient
\end{tabular} \\
\hline
\rowcolor{thirdcolor}
\multicolumn{5}{l}{\textbf{Cyclicality}} \\
\rowcolor{primarycolor}
& \textbf{Content} &
Does the output contain content‑related or literal repetitions? &
\begin{tabular}{lp{5.7cm}}
5 & no repetitions (content‑related or literal) \\
4 &  \\
3 & one repetition (content‑related or literal) \\
2 &  \\
1 & quite a lot of repetitions (content‑related or literal)
\end{tabular} \\
\hline
\end{longtable}
\section{Discussion}
\subsection{Findings and Relation to Previous Research}
In this paper, we presented a human-centred questionnaire aimed at evaluating RAG outputs. Overall, the questionnaire was well-received by raters, with only minor suggestions for adjustments. While LLM ratings were generally perceived as useful and logical, it was noted that they were only partially correct and inconsistent with the rater's evaluations. The LLM demonstrated a superior ability compared to humans to focus on the aspects specified in the metric description and the scale labels. However, it demonstrated certain limitations in detecting different format styles. This finding indicates that questionnaires must be worded consistently and clearly to ensure that they are used correctly by human raters. LLMs can provide additional support by strictly focusing on the given information and enhancing ratings with explanations.
\par
The utility dimensions proposed by \cite{gienappEvaluatingGenerativeAd2024} provided an appropriate groundwork for a human-centered questionnaire. However, several drawbacks were identified:
\begin{enumerate}
    \item We decided not to include "factual correctness" in our final questionnaire. As "external consistency" already evaluates the relationship between the source document and the output, it might be sufficient in a RAG context, regardless of the objective truth of the information in the source document. When evaluating statements, their verifiability is a key consideration.
    \item The output must be presented logically and consistently. However, in addition, it should be accompanied by effective formatting strategies to enhance user comprehension.
    \item In addition to being logical, effective text structuring is also important. It is essential to avoid repetition and to ensure that key information is clearly and prominently featured in the text.
    \item Most metrics must be viewed in the context of the intent displayed in the user query. For instance, if users employ operators such as "summarise", they may expect a different output than "explain" or "elaborate".
    \end{enumerate}
In comparison to computer-centred frameworks, this questionnaire is similar in that it also incorporates accuracy-related metrics such as "external consistency" and "language correctness". However, many of the metrics that have been introduced have a focus on enhancing text readability, with the aim of improving human understanding. Furthermore, metrics are either evaluated by a human or a human-LLM collaboration, but never numerically calculated. In comparison to generic LLM dialogue evaluation frameworks, such as the one by \cite{ikeAutomatingDialogueEvaluation2025}, metrics were focused on the alignment of the query-output pairs with the given source document. For instance, "commonsense" is not necessary in a RAG context, since outputs only need to make sense in the context of the given source document. Another example is "innovation", which involves creative outputs that are not adequate for a RAG context. However, "redundancy" resembles the "content cycliality" metric, which we added to the questionnaire.
\subsection{Ethical Implications}
When evaluating LLM outputs with other LLMs, there are important ethical considerations. If machines are to be evaluated by machines, how can we ensure that outputs are understandable or useful to human users? On the other hand, if we only consider a human-centered perspective, how can we still ensure that systems are accurate and performant?
\par
To perceive human agency, it is important to include human users and human-centered metrics in the evaluation process. In the present work, query-output pairs were evaluated by a human and a human-LLM collaboration. In the latter pair, the final rating decision was made by the human. The LLM rating and explanations were solely used to enhance the human's cognitive process while evaluating the outputs. 
There is still a risk of the LLM "misunderstanding" the task and negatively influencing the human. However, this would also be the case if humans collaboratively rated RAG outputs. Furthermore, metrics that compare the output to the statements in the source documents should not be evaluated by an LLM. That is, because during the RAG process, this comparison has already been made, and employing another LLM might further diminish the accuracy of the results. Even though "factual correctness" is not present in the questionnaire anymore, it is important to consider that this is an essential metric in non-RAG contexts. When questioning source documents, "truth" is only valid in the context of the given information. To avoid the risk of misinformation, factually correct source documents should be chosen. However, assessing the quality of the source documents is a different matter.
Suggestions for RAG evaluations are the following:
\begin{enumerate}
    \item The final decision for a numeric rating should stay with the human.
    \item It is recommended that LLM ratings comprise not only numeric ratings and explanations, but also direct examples from the source text and outputs for better verification.
    \item There should be a balance between the employment of human-centered metrics and computer-centered metrics.
    \item Metrics evaluating the contextual "truth" (e.g., consistency between source document and output) should be evaluated solely by the human. 
\end{enumerate}
\subsection{Practical Implications}
The questionnaire can be readily incorporated into RAG evaluation processes due to its comprehensive metric descriptions and scale labels. For instance, it can be used to evaluate query-output pairs following user testing or simply to optimize the RAG pipeline and prompts beforehand. On the other hand, the necessary time required for the evaluation process must be considered. Raters must be familiar with the evaluation metrics and the source documents before conducting the evaluation. Furthermore, we recommend that at least two human raters are involved to reduce bias and increase the precision of the ratings. The generation of LLM tokens presents another cost factor, which varies according to the employed LLM. 
For these reasons, ratings should be conducted for a selected subset of query-output pairs. For instance, queries could be categorized based on operators ("summarize", "name", "elaborate", etc.), or the number of documents involved (specific requests concerning one document vs. generic requests concerning multiple documents). Subsequently, each category can be evaluated with a query-output pair. 
\subsection{Limitations}
The questionnaire does not come without limitations. Firstly, the emphasis is on textual query-output pairs. This approach is limited in scope as it excludes systems capable of processing visual inputs and generating visual outputs. However, evaluating visual outputs would involve a very different set of metrics, as textual and visual information are processed differently by humans \cite{baddeleyEpisodicBufferNew2000}.
Another limitation concerns the validity of the questionnaire. Despite the fact that two raters evaluated it, the small number of human evaluators is not sufficient to ensure questionnaire validity. Single ratings may be subject to bias and lack reliability. Therefore, numeric results are not generalizable. 
A third limitation lies in the application of the questionnaire. As previously mentioned, human ratings are valuable, but their generation is effort-and time-consuming. 
Finally, query-output pairs that were generated in the user tests are limited by their scope and are influenced by the tasks conducted during the user tests. When evaluating different queries, other metrics might be important in addition.
\subsection{Future Research}
Inherently, "model clarity" was introduced as a metric in the questionnaire to assess the system's explainability. However, ultimately, we excluded it because giving an understandable textual explanation for every output that exceeds simple text references might not be user-friendly. In addition to textually referencing parts of the source document, "token highlighting" is a visual explainability method that can be employed in RAG systems \cite{wangRAGVizDiagnoseVisualize2024}. Since explainability is important to heighten perceived transparency and calibrate user trust \cite{kamathExplainableArtificialIntelligence2021}, research could incorporate metrics aimed at evaluating common explainability methods in RAG systems.
\par
In the current study, only separated query-output pairs have been examined. However, to humans, experiencing human-LLM interaction as a social dialogue is an important concern \cite{millerExplanationArtificialIntelligence2019}. Evaluating the dialogue as a whole might shed more light on the perceived fluency of the conversation. For instance, \cite{borsciChatbotUsabilityScale2022}  provide several items regarding dialogue flow in their "Chatbot Usability Scale". 
In addition to dialogue flow, the current framework could be extended by adding relevant computer-centered metrics. It is important to balance human-centeredness and computer-centeredness to gain a holistic view of a system.
\par
To ensure the questionnaire’s psychometric soundness, it should be subjected to a formal validation process that incorporates both a sufficiently large rater pool and input from domain‑specific experts. Reliability can be quantified using internal‑consistency metrics such as Cronbach’s Alpha, which provides an estimate of the extent to which items measure a common construct \cite{moosbruggerTesttheorieUndFragebogenkonstruktion2020}. Content validity, in turn, is best judged by subject‑matter specialists who can evaluate whether the items comprehensively represent the intended construct. Accordingly, user experience specialists and RAG experts, who possess the requisite knowledge of the questionnaire’s target domain, should review item relevance, clarity, and coverage. Conducting these assessments with an adequately sized sample of raters can provide reliability and content validity estimates and generate suggestions for improvement.
\section{Conclusion}
This paper presents a human-centred questionnaire based on \cite{gienappEvaluatingGenerativeAd2024} framework, which aids the evaluation of RAG outputs. The development process involved repeatedly applying the metrics to given query-output pairs, as well as discussions and refinements regarding precise item formulation and scale labeling. The initial questionnaire was evaluated by two raters and received favourable feedback. Following the implementation of raters' suggestions, we created the final questionnaire, extending the initial framework by emphasizing the match between output and user intent, structuring text content, and avoiding repetitions. A key change was the decision to evaluate factual truth solely in the context of source documents.
Limitations include the lack of a validation study involving an adequate number of raters and the focus on textual outputs only.
Future research should validate, extend, and apply the questionnaire in different contexts.
Overall, the questionnaire provides a solid foundation for further exploration and can be incorporated into the system development and evaluation process.

	\clearpage
    \printbibliography
	\appendix
    \end{document}